\documentclass{jpsj3}
\usepackage{txfonts}
\usepackage{bm}
\usepackage{graphicx}
\usepackage{color}
\usepackage{float}

\title{Performance Evaluation of Ising and QUBO Variable Encodings in Boltzmann Machine Learning}

\author{Yasushi Hasegawa$^{1,2}$\thanks{hasegawa.yasushi.s6@dc.tohoku.ac.jp}, and Masayuki Ohzeki$^{1,3,4,5}$}
\inst{$^1$Graduate School of Information Sciences, Tohoku University, Sendai 980-8579, Japan \\
$^2$Technology Strategy Division, SCSK Corporation, Koto, Tokyo 135-8110, Japan \\
$^3$Department of Physics, Institute of Science Tokyo, Meguro, Tokyo 152-8551, Japan \\
$^4$Research and Education Institute for Semiconductors and Informatics, Kumamoto University, Chuo, Kumamoto 860-0862, Japan \\
$^5$Sigma-i Co., Ltd., Minato, Tokyo 108-0075, Japan} 

\abst{
We compare Ising ($\{-1,+1\}$) and QUBO ($\{0,1\}$) encodings for Boltzmann machine learning under controlled protocols that fix the sampler, optimizer, and learning-rate design within each comparison. Exploiting the identity that the Fisher information matrix (FIM) equals the covariance of sufficient statistics, we visualize empirical moments from model samples and reveal systematic, representation-dependent differences. QUBO induces larger cross terms between first- and second-order statistics, creating more small-eigenvalue directions in the FIM and lowering spectral entropy. This ill-conditioning explains slower convergence under stochastic gradient descent (SGD). 
In contrast, full-FIM natural gradient descent (NGD), which rescales updates by the FIM metric, achieves similar convergence across encodings, whereas diagonal-FIM approximation can reintroduce representation-dependent differences. Practically, for SGD-based training, the Ising encoding provides more isotropic curvature and faster convergence; for QUBO, centering/scaling or NGD-style preconditioning mitigates curvature pathologies. These results clarify how representation shapes information geometry and finite-time learning dynamics in Boltzmann machines and yield actionable guidelines for variable encoding and preprocessing.
}

\begin{document}
\maketitle

\section{Introduction}

Binary systems in combinatorial optimization and energy-based learning are commonly formulated as the Ising model ($\boldsymbol{s} \in \{-1, +1\}$) or the Quadratic Unconstrained Binary Optimization (QUBO) model ($\boldsymbol{x} \in \{0, 1\}$). The two encodings are related by the affine transformation $s = 2x - 1$ and are theoretically equivalent to their optimal solutions~\cite{lucas2014ising, glover2022quantum}. With appropriate parameter mapping, either representation can reach the same ground state of a given problem.

However, in practical optimization and learning dynamics, this equivalence breaks down. Variable encoding, scaling, and penalty design differences can lead to markedly distinct convergence behaviors and robustness. For example, studies on the Fujitsu Digital Annealer have shown that solving the same problem using Ising and QUBO formulations can yield smoother and more stable trajectories under the Ising representation~\cite{kaseb2024power}. Similarly, comparing QUBO transformation schemes for 3-SAT demonstrates that the transformation significantly affects solution quality on identical hardware~\cite{munch2023transformation}. In Ising-machine-based solvers for linear systems, the choice of real-number encoding also influences accuracy and error characteristics, reflecting differences in search geometry and the ease of crossing energy barriers~\cite{endo2024novel}. These observations collectively indicate that even for the same logical problem, the coordinate system—i.e., the representation—can strongly affect finite-time optimization behavior.

This issue naturally extends to Boltzmann Machine (BM) learning~\cite{ackley1985learning}, where iterative, sampling-based training methods such as Contrastive Divergence rely on accurate gradient estimates. The distribution output by a quantum annealer, due to freeze-out effects, can often be modeled as a Gibbs–Boltzmann form; hence, annealers have been used as physical samplers for BM and Restricted Boltzmann Machine (RBM) learning~\cite{sato2021assessment, hasegawa2026kernel, urushibata2022comparing}. In classical and quantum settings, gradient scale and data centering influence convergence through conditioning and gradient variance~\cite{montavon2012deep, hinton2012practical, melchior2016center}. The Ising encoding naturally offers zero-mean centering and symmetric scaling of interaction terms, which can reduce bias and variance in gradient estimates and facilitate step-size tuning. In contrast, the QUBO encoding lacks this centering property. From an information-geometric perspective, switching from QUBO to Ising effectively performs centering and rescaling equivalent to preconditioning by the Fisher information matrix (FIM)~\cite{amari1998natural, martens2015optimizing}.

Natural Gradient Descent (NGD) addresses curvature anisotropy by rescaling updates using the FIM metric. Both theoretical and empirical studies show that NGD accelerates and stabilizes learning compared with first-order methods~\cite{amari1998natural, melchior2016center, desjardins2013metric}. Investigating how encoding-dependent changes in the FIM spectrum affect convergence thus provides a principled way to understand the influence of variable representation on learning dynamics.

This study systematically examines how variable encoding—Ising versus QUBO—affects convergence and conditioning in BM learning under strictly controlled conditions, fixing the model, sampler, and learning-rate design. We analyze the relationship between encoding and the information-geometric curvature characterized by FIM spectra and spectral entropy. By comparing Stochastic Gradient Descent (SGD) and Natural Gradient Descent (NGD), we clarify how encoding and learning rules interact to determine convergence speed and stability. Our findings yield practical insights into representation choice and preprocessing strategies for efficient BM learning. In this paper, ``preprocessing'' refers specifically to centering and scaling of variables or sufficient statistics induced by the affine transformation between
QUBO and Ising encodings, rather than to generic data preprocessing such as feature normalization of the input dataset.

The following section introduces the Ising and QUBO formulations, the BM model, and the SGD and NGD update rules. The Results section presents learning trajectories and FIM spectral analyses for both encodings. Finally, the Conclusion summarizes the key findings and outlines future research directions.

\section{Methods}

\subsection{Model and Representations}
In the main experiments, we consider the same learning model, namely a fully connected BM, the same sampler (Simulated Annealing (SA)), and the same update rules (SGD/NGD), and vary only the variable representation: Ising $\{-1,+1\}$ versus QUBO $\{0,1\}$. Additional RBM experiments are used to examine whether the observed behavior also appears in a sparse architecture.
The Ising energy is
\begin{equation}
E_{\mathrm{Ising}}(\boldsymbol{s};\boldsymbol{h},\boldsymbol{J})
= \sum_i h_i s_i + \sum_{i<j} J_{ij} s_i s_j.
\end{equation}
The QUBO energy is
\begin{equation}
E_{\mathrm{QUBO}}(\boldsymbol{x};\boldsymbol{Q})
= \sum_{i\le j} Q_{ij} x_i x_j.
\end{equation}
The two representations are related by $ s=2x-1 $, which gives the following parameter correspondence (constant terms omitted):
\begin{equation}
h_i=\tfrac14\sum_j (Q_{ij}+Q_{ji}),\qquad J_{ij}=\tfrac14 Q_{ij}.
\end{equation}

\subsection{Boltzmann Machine}
At temperature $ T $ (inverse temperature $ \beta=1/T $), the Gibbs–Boltzmann distribution is
\begin{equation}
P_\theta(\boldsymbol{s})=\frac{1}{Z(\theta)}\exp\{-\beta E_{\mathrm{Ising}}(\boldsymbol{s};\theta)\},
\end{equation}
\begin{equation}
Z(\theta)=\sum_{\boldsymbol{s}}\exp\{-\beta E_{\mathrm{Ising}}(\boldsymbol{s};\theta)\},
\end{equation}
and analogously for QUBO.
The average negative log‑likelihood $ L( \theta ) = -\frac{1}{N} \sum_{n=1}^{N} \log P_{\theta} (\boldsymbol{s}^{n}) $ has gradient
\begin{equation}
\nabla L(\theta)=\beta\big(\mathbb{E}_{\text{data}}[\nabla_\theta E]-\mathbb{E}_{P_\theta}[\nabla_\theta E]\big).
\end{equation}
Componentwise, $ \nabla_{h_{i}} E=s_{i} $, $ \nabla_{J_{ij}} E=s_i s_j $, and $ \nabla_{Q_{ij}} E=x_i x_j $.
The model expectation $ \mathbb{E}_{P_{\theta}}[\cdot] $ is approximated by SA with single spin flips and Metropolis acceptance $ \alpha = \min \{1,\exp(-\beta \Delta E)\} $.

\subsection{Stochastic Gradient Descent (SGD) and Natural Gradient Descent (NGD)}
A second-order Taylor approximation near $ \theta_t $ gives
\begin{equation}
L(\theta_t+\delta)\approx L(\theta_t)+\nabla L(\theta_t)^\top\delta+\tfrac12\,\delta^\top \nabla^2 L(\theta_t)\,\delta.
\end{equation}
At a local minimizer, $ \nabla L(\theta_t)+\nabla^2 L(\theta_t)\delta=0 $, hence $ \delta = -(\nabla^{2} L)^{-1} \nabla L $.
SGD updates
\begin{equation}
\theta_{t+1}=\theta_t-\eta_{\mathrm{SGD}}\nabla L(\theta_t),
\end{equation}
with a stability condition \cite{horn2012matrix} $ \rho(I-\eta\nabla^{2} L)<1 $, namely $ 0<\eta<2/\lambda_{\max}(\nabla^2L) $.
NGD updates
\begin{equation}
\theta_{t+1}=\theta_t-\eta_{\mathrm{NGD}}\,(\nabla^2L(\theta_t))^{-1}\nabla L(\theta_t).
\end{equation}

The Hessian equals the FIM for Boltzmann distributions. 
In fact
\begin{equation}
\mathbb{E}\!\left[\frac{\partial^2 \log P}{\partial\theta_i\partial\theta_j}\right]
=-\mathbb{E}\!\left[\frac{\partial \log P}{\partial\theta_i}\frac{\partial \log P}{\partial\theta_j}\right]
=-F_\theta.
\end{equation}
We thus employ a damped NGD step
\begin{equation}
\theta_{t+1}=\theta_t-\eta_{\mathrm{NGD}}\,(F_\theta+\lambda I)^{-1}\nabla L(\theta_t).
\end{equation}

Moreover, for Boltzmann distributions, the FIM equals the covariance matrix of sufficient statistics. For QUBO parameters $ Q_{ij} $:
\begin{equation}
\mathbb{E}\!\left[\frac{\partial \log P}{\partial Q_{ij}}\frac{\partial \log P}{\partial Q_{kl}}\right]
=\mathbb{E}\big[(x_ix_j-\mathbb{E}[x_ix_j])(x_kx_l-\mathbb{E}[x_kx_l])\big]
=\mathbb{E}[x_ix_jx_kx_l]-\mathbb{E}[x_ix_j]\mathbb{E}[x_kx_l].
\end{equation}
For Ising parameters $ h_i $, $ J_{ij} $:
\begin{align}
\mathbb{E} \!\left[ \frac{\partial \log P}{\partial h_i}\frac{\partial \log P}{\partial h_j}\right]
&=\mathbb{E}[s_is_j]-\mathbb{E}[s_i]\mathbb{E}[s_j],\\
\mathbb{E} \!\left[ \frac{\partial \log P}{\partial h_i} \frac{\partial \log P}{\partial J_{kl}} \right]
&=\mathbb{E}[s_is_ks_l]-\mathbb{E}[s_i]\mathbb{E}[s_ks_l],\\
\mathbb{E} \!\left[ \frac{\partial \log P}{\partial J_{ij}} \frac{\partial \log P}{\partial J_{kl}} \right]
&=\mathbb{E}[s_is_js_ks_l]-\mathbb{E}[s_is_j]\mathbb{E}[s_ks_l].
\end{align}

\subsection{Spectral Entropy}
Let $ \{ \lambda_i \} $ be the FIM eigenvalues. The spectral entropy is
\begin{equation}
S=-\sum_i \frac{\lambda_i}{\sum_j \lambda_j}\log\!\left(\frac{\lambda_i}{\sum_j \lambda_j}\right),
\end{equation}
which measures the evenness of curvature allocation, attaining its maximum when all eigenvalues are equal.

\subsection{Restricted Boltzmann Machine}
We additionally performed RBM experiments to examine whether the representation-dependent learning behavior observed in the fully connected BM also appears in a sparse architecture. Let $v_i$ and $h_j$ denote visible and hidden variables, respectively, where $i=1,\ldots,n_v$ and $j=1,\ldots,n_h$. For the Ising encoding, $v_i,h_j\in\{-1,+1\}$, and the RBM energy was defined as
\begin{equation}
E_{\mathrm{RBM}}^{\mathrm{Ising}}(\boldsymbol{v},\boldsymbol{h}) = \sum_i b_i v_i + \sum_j c_j h_j + \sum_{i,j} W_{ij}v_i h_j .
\end{equation}
For the QUBO encoding, $x_i,y_j\in\{0,1\}$, and the corresponding RBM energy was defined as
\begin{equation}
E_{\mathrm{RBM}}^{\mathrm{QUBO}}(\boldsymbol{x},\boldsymbol{y}) = \sum_i b^{\mathrm{QUBO}}_i x_i + \sum_j c^{\mathrm{QUBO}}_j y_j + \sum_{i,j} W^{\mathrm{QUBO}}_{ij}x_i y_j .
\end{equation}
Thus, the parameter vectors are $\theta_{\mathrm{Ising}}=(\boldsymbol{b},\boldsymbol{c},W)$ and $\theta_{\mathrm{QUBO}}=(\boldsymbol{b}^{\mathrm{QUBO}},\boldsymbol{c}^{\mathrm{QUBO}},W^{\mathrm{QUBO}})$. The two encodings are related by $x_i=(v_i+1)/2$ and $y_j=(h_j+1)/2$. In the implementation, the initial parameters were first randomly generated in the QUBO parameterization and then converted to the Ising parameterization as
\begin{align}
b_i &= \frac{1}{2}b^{\mathrm{QUBO}}_i + \frac{1}{4}\sum_j W^{\mathrm{QUBO}}_{ij},\\
c_j &= \frac{1}{2}c^{\mathrm{QUBO}}_j + \frac{1}{4}\sum_i W^{\mathrm{QUBO}}_{ij},\\
W_{ij} &= \frac{1}{4}W^{\mathrm{QUBO}}_{ij},
\end{align}
up to an additive constant in the energy. The QUBO initial parameters were sampled from a normal distribution with standard deviation $0.1$.

The sufficient statistics used for the gradients and the FIM were $(v_i,h_j,v_i h_j)$ for the Ising RBM and $(x_i,y_j,x_i y_j)$ for the QUBO RBM. For the data-dependent term, the visible variables were fixed to the training data, and the hidden-variable expectations were computed analytically using the conditional distribution. With the sign convention of the energy above, the Ising conditional expectation was computed as
\begin{equation}
\langle h_j\rangle_{\boldsymbol{v}} = -\tanh\left[\beta \left(c_j+\sum_i W_{ij}v_i \right)\right].
\end{equation}
Equivalently, this was implemented through $p(h_j=+1|\boldsymbol{v})=1/[1+\exp\{2\beta(c_j+\sum_i W_{ij}v_i)\}]$ and $\langle h_j\rangle_{\boldsymbol{v}}=2p(h_j=+1|\boldsymbol{v})-1$. The data-dependent statistics were then evaluated as $\langle h_j\rangle_{\mathrm{data}}=\mathbb{E}_{\mathrm{data}}[\langle h_j\rangle_{\boldsymbol{v}}]$ and $\langle v_i h_j\rangle_{\mathrm{data}}=\mathbb{E}_{\mathrm{data}}[v_i\langle h_j\rangle_{\boldsymbol{v}}]$. For the QUBO RBM, the conditional expectation was computed as
\begin{equation}
\langle y_j\rangle_{\boldsymbol{x}} = \frac{1}{1+\exp\left[\beta \left( c^{\mathrm{QUBO}}_j+\sum_i W^{\mathrm{QUBO}}_{ij}x_i \right) \right]}.
\end{equation}
The corresponding data-dependent statistics $\langle y_j\rangle_{\mathrm{data}}$ and $\langle x_i y_j\rangle_{\mathrm{data}}$ were computed in the same manner.

The model-dependent term was estimated from joint SA samples of visible and hidden variables. For the Ising RBM, the RBM parameters were converted to an Ising Hamiltonian over $n_v+n_h$ spins, and for the QUBO RBM they were converted to a QUBO matrix over $n_v+n_h$ binary variables. The resulting joint samples were used to estimate the model-dependent expectations of the same sufficient statistics used in the data-dependent term.

The FIM was computed as the covariance matrix of the sufficient-statistic vector under the model distribution,
\begin{equation}
T_{\mathrm{RBM}} = (\boldsymbol{v},\boldsymbol{h},\mathrm{vec}(\boldsymbol{v}\boldsymbol{h}^{\mathrm{T}}))
\end{equation}
for the Ising RBM, and analogously for the QUBO RBM. The covariance was estimated from the same joint model samples used for the model-dependent gradient term. For numerical stability, the FIM was symmetrized and damped by adding $10^{-3}I$. Eigenvalues smaller than $10^{-3}$ were thresholded only when storing the FIM spectrum, estimating the effective rank, and computing the normalized SGD learning rate. The NGD update used the Moore--Penrose pseudoinverse of the damped FIM.

The KL divergence reported for the RBM experiments was evaluated on the visible-variable distribution. Specifically, the model visible distribution was estimated from the empirical histogram of the visible components of the joint model samples.

\section{Results}
\subsection{Experimental setup}
We compare Ising and QUBO encodings while fixing the sampler, optimizer, and learning-rate design within each comparison. We conducted two classes of experiments: fully connected BM experiments, which constitute the main setting of this study, and additional RBM experiments, which examine whether the same qualitative behavior also appears in a sparse architecture.

For the fully connected BM experiments, the datasets were Bars and Stripes (BAS) $2\times2$ with $450$ samples, BAS $3\times3$ with $1{,}120$ samples, BAS $4\times4$ with $2{,}590$ samples, and Ising-sampled synthetic data with dimension $d=10$ and couplings $J_{ij}\sim \mathcal{N}(0,J_c^2/d)$, where $J_c\in\{0.5,1.0,1.5\}$. For the RBM experiments, we used the BAS $2\times2$ and BAS $3\times3$ datasets and fixed the number of hidden units to $n_h=4$. The RBM formulation, parameter conversion, sufficient statistics, and expectation computations are described in the Methods section.

For both the fully connected BM and RBM experiments, we used simulated annealing (SA) with $\beta=1.0$ and $10{,}000$ model samples at each iteration. In the RBM experiments, these samples correspond to joint visible--hidden samples. SGD and NGD were performed under the same learning-rate design. For SGD, the learning rate was set to $\eta_{\mathrm{SGD}}=0.01/\lambda_{\max}(F_{\theta})$, unless otherwise stated. For NGD, we used $\eta_{\mathrm{NGD}}=0.01$ with damping $\lambda=10^{-3}$.

\subsection{Learning trajectories in fully connected BMs}
\begin{figure}[htbp]
  \centering
  \includegraphics[width=1.0\linewidth]{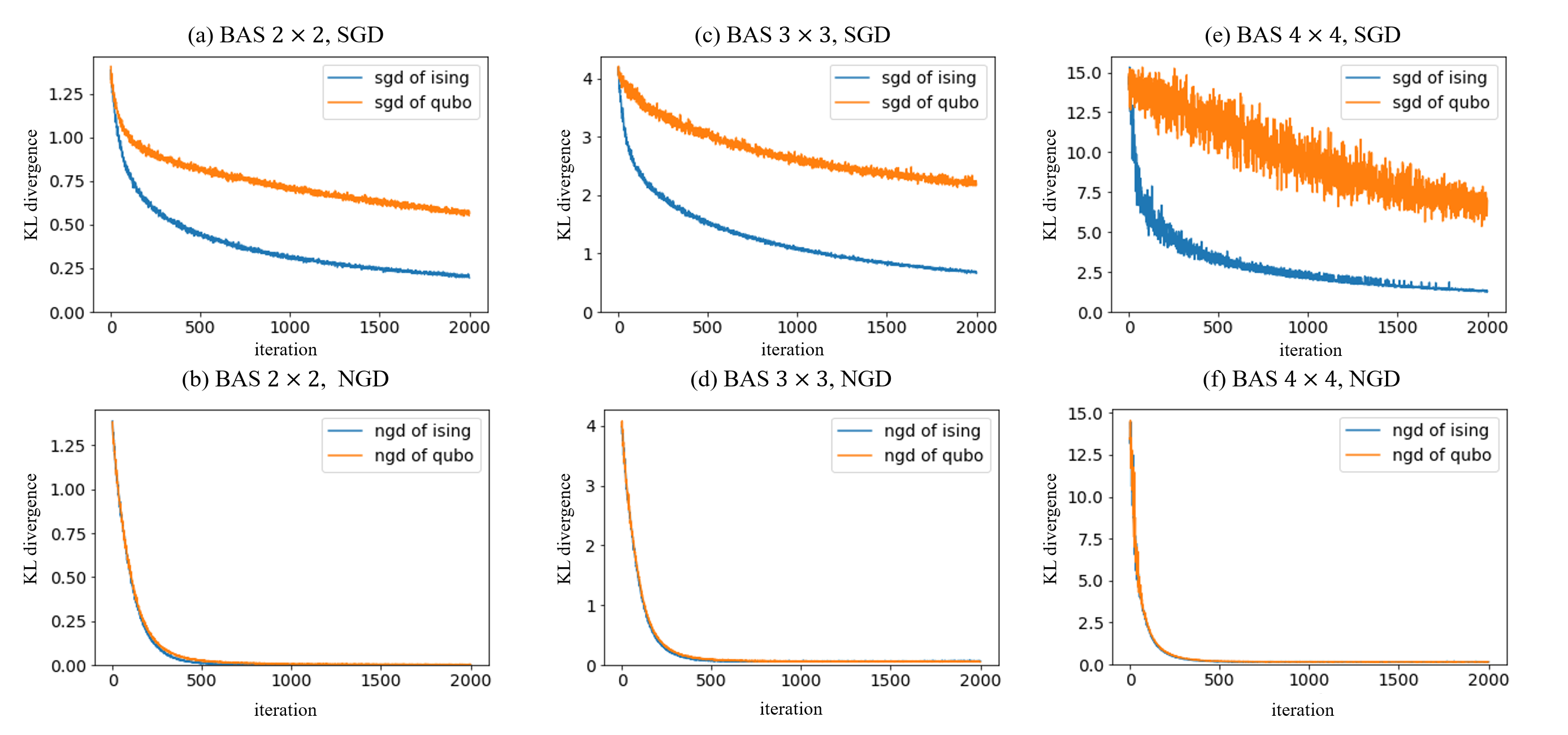}
  \caption{(Color online) Learning trajectories on BAS datasets using the fully connected BM setting. Panels (a), (c), and (e) show SGD results for BAS $2\times2$, BAS $3\times3$, and BAS $4\times4$, respectively, while panels (b), (d), and (f) show the corresponding NGD results. The numbers of training samples were $450$, $1{,}120$, and $2{,}590$ for BAS $2\times2$, BAS $3\times3$, and BAS $4\times4$, respectively. Under SGD, the Ising encoding reduces the KL divergence faster than the QUBO encoding for all BAS sizes, including the larger BAS $4\times4$ case. In contrast, under NGD, the convergence trajectories of the two encodings are nearly identical, indicating that FIM-based preconditioning compensates for the representation-dependent curvature difference.}
  \label{fig:kls_bas_fullbm}
\end{figure}

Fig. \ref{fig:kls_bas_fullbm} compares the KL-divergence trajectories on BAS datasets using the fully connected BM setting. In panels (a), (c), and (e) of Fig. \ref{fig:kls_bas_fullbm}, SGD is used for BAS $2\times2$, BAS $3\times3$, and BAS $4\times4$, respectively. In all three cases, the Ising encoding decreases the KL divergence faster than the QUBO encoding. The same qualitative tendency is observed even for BAS $4\times4$, which has a larger visible dimension and $2{,}590$ training samples. This result indicates that the observed advantage of Ising under SGD is not restricted to the smallest BAS datasets. In contrast, panels (b), (d), and (f) of Fig. \ref{fig:kls_bas_fullbm} show that NGD yields nearly identical convergence trajectories for the two encodings. Thus, when the FIM metric is explicitly used to compensate for local curvature, the representation-dependent difference is strongly reduced.
These trajectory results show that the Ising encoding has a clear convergence advantage under SGD, whereas the use of the full FIM metric in NGD substantially reduces the representation-dependent gap.

\subsection{Fixed-learning-rate SGD and FIM eigenvalues during SGD}
\begin{figure}[htbp]
  \centering
  \includegraphics[width=1.0\linewidth]{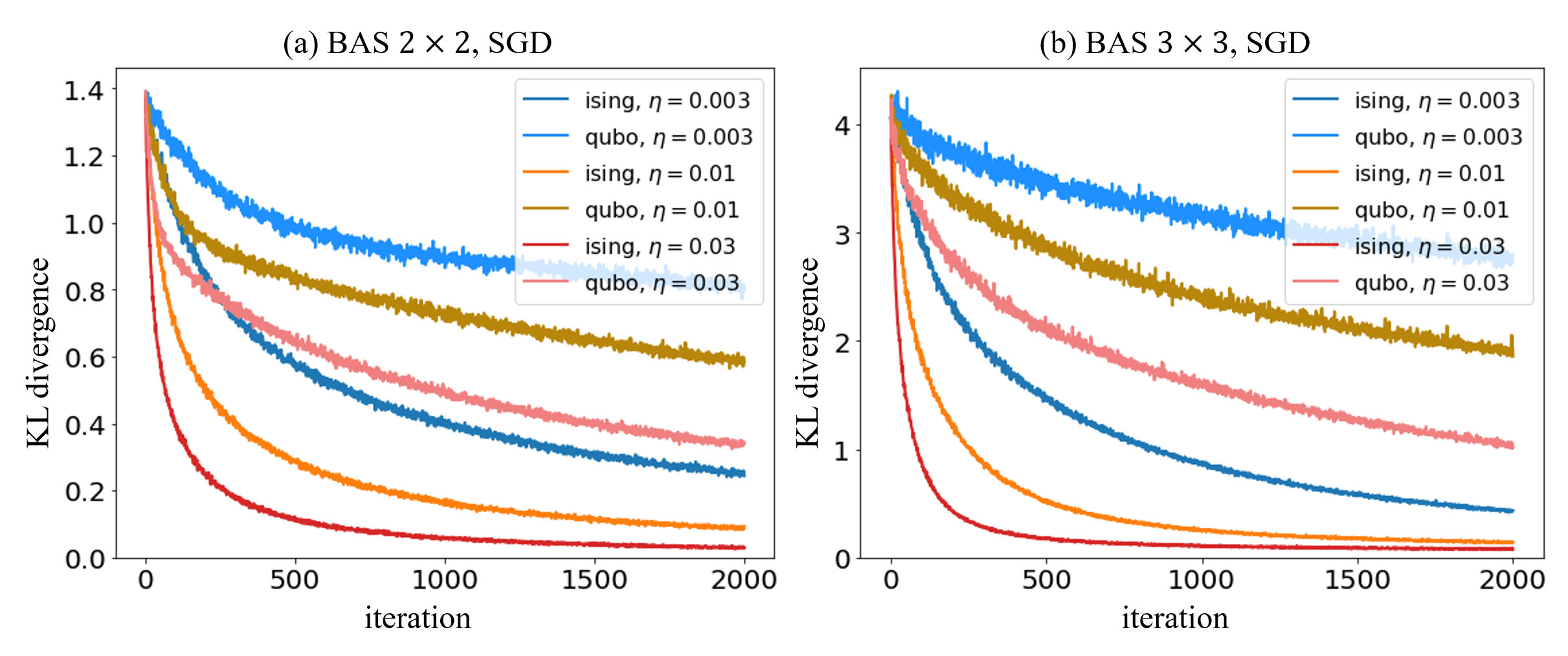}
  \caption{(Color online) Fixed-learning-rate SGD learning trajectories on BAS datasets without normalization by $\lambda_{\max}(F_{\theta})$. Panels (a) and (b) show the results for BAS $2\times2$ and BAS $3\times3$, respectively. For each dataset, the same fixed learning rates, $\eta=0.003$, $0.01$, and $0.03$, were used for both the Ising and QUBO encodings. Across the tested learning rates, the Ising encoding reduces the KL divergence faster than the QUBO encoding. These results indicate that the observed SGD advantage of the Ising encoding is not specific to $\eta=0.01$ and is not solely an artifact of the $\lambda_{\max}(F_{\theta})$-normalized learning-rate rule.}
  \label{fig:kls_bas_lr_fixed}
\end{figure}

\begin{figure}[htbp]
  \centering
  \includegraphics[width=1.0\linewidth]{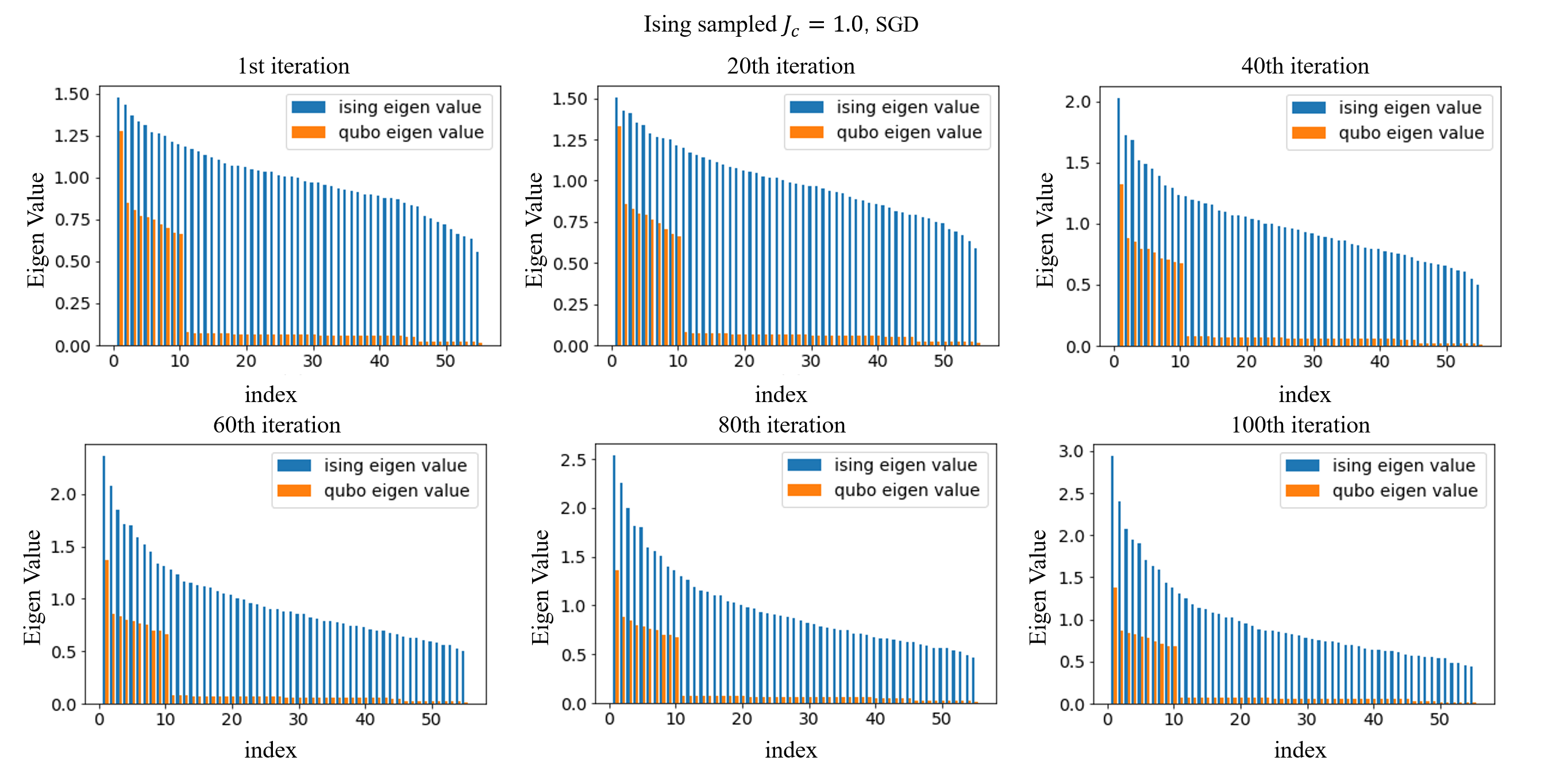}
  \caption{(Color online) Evolution of FIM eigenvalues during SGD on Ising-sampled data with $J_c=1.0$. The eigenvalues are sorted in descending order at each iteration. The maximum eigenvalue of the Ising FIM is larger than that of the QUBO FIM throughout the plotted early training stage. Therefore, under the normalized learning-rate rule $\eta_{\mathrm{SGD}}=0.01/\lambda_{\max}(F_{\theta})$, the effective learning rate assigned to QUBO is larger than that assigned to Ising. Nevertheless, the KL-divergence results show faster convergence for Ising, suggesting that the slower convergence of QUBO is not caused by an unfairly small learning rate, but rather by its anisotropic FIM spectrum with many small-eigenvalue directions.}
  \label{fig:eigenvals_ising_sgd}
\end{figure}

Fig. \ref{fig:kls_bas_lr_fixed} further examines whether the SGD result depends on the $\lambda_{\max}(F_{\theta})$-normalized learning-rate rule or on a particular choice of the fixed learning rate. In this experiment, fixed learning rates $\eta=0.003$, $0.01$, and $0.03$ were directly applied to both encodings without normalization by $\lambda_{\max}(F_{\theta})$. Panels (a) and (b) show the results for BAS $2\times2$ and BAS $3\times3$, respectively. For all tested learning rates and both datasets, the Ising encoding decreases the KL divergence faster than the QUBO encoding. Although the convergence speed depends on the learning rate, the qualitative tendency that the Ising encoding is more favorable for SGD remains unchanged over the tested range. Therefore, the faster convergence of Ising under SGD is not specific to $\eta=0.01$ and is not solely caused by the normalized learning-rate rule.

Fig. \ref{fig:eigenvals_ising_sgd} provides an additional check on the learning-rate normalization. The maximum eigenvalue of the Ising FIM is larger than that of the QUBO FIM during the plotted early training stage. Therefore, under the normalized learning-rate rule $\eta_{\mathrm{SGD}}=0.01/\lambda_{\max}(F_{\theta})$, the effective learning rate assigned to QUBO is larger than that assigned to Ising. Nevertheless, the KL-divergence results show faster convergence for the Ising encoding. This result indicates that the slower convergence of QUBO is not caused by an unfairly small learning rate, but is instead consistent with its anisotropic FIM spectrum and many small-eigenvalue directions.

\subsection{Synthetic Ising-sampled data and FIM spectral structure}
\begin{figure}[htbp]
  \centering
  \includegraphics[width=1.0\linewidth]{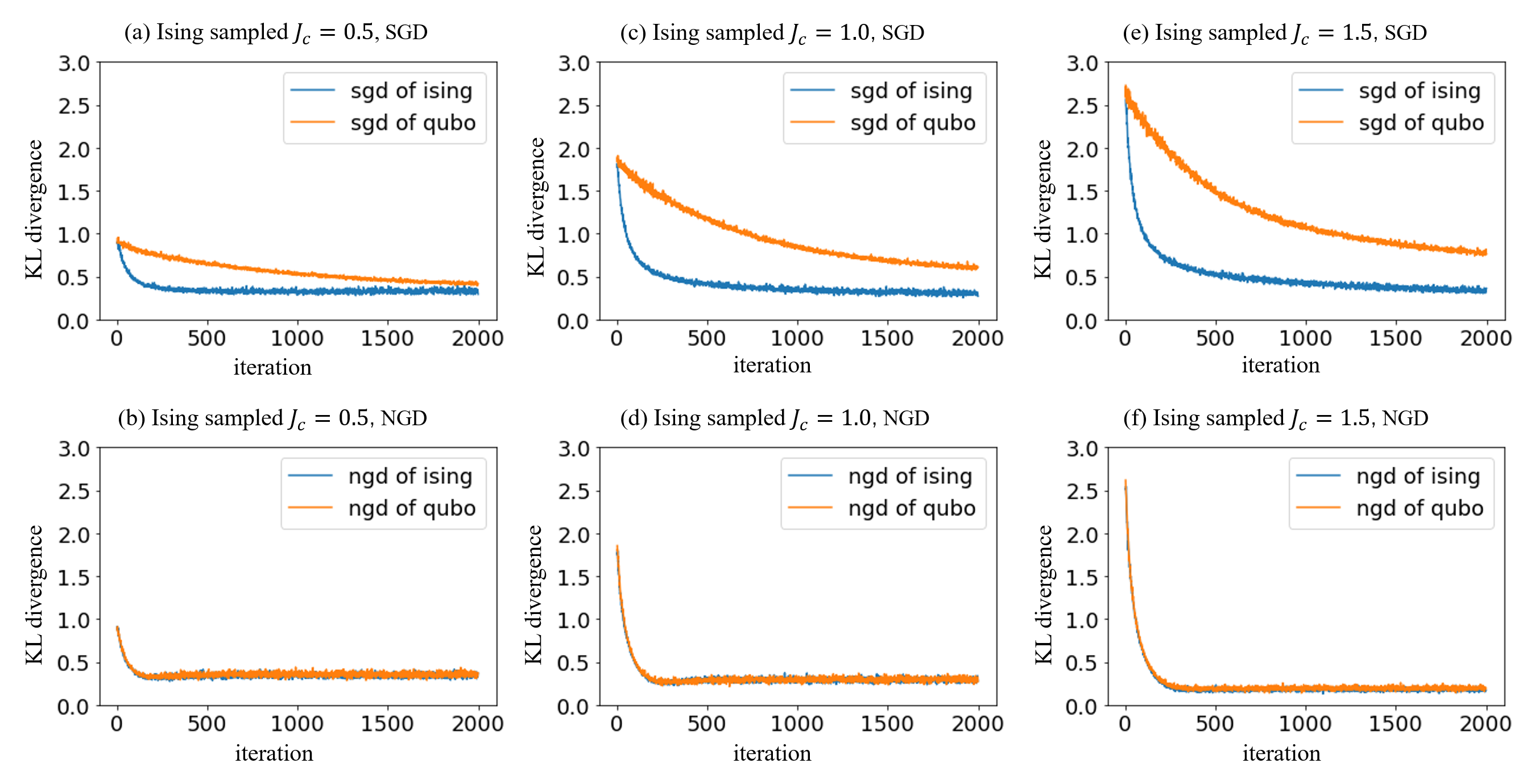}
  \caption{(Color online) Learning trajectories on Ising-sampled synthetic data with $d=10$ for $J_c=0.5$, $1.0$, and $1.5$. Panels (a), (c), and (e) show SGD results, whereas panels (b), (d), and (f) show NGD results. Under SGD, the Ising encoding converges faster than the QUBO encoding for all values of $J_c$, and the gap becomes more visible as the coupling scale increases. Under NGD, the trajectories of the two encodings are nearly identical for all values of $J_c$. These results show that the encoding-dependent convergence difference is not limited to BAS datasets and is also observed in synthetic data generated from Ising-type interactions.}
  \label{fig:kls_ising}
\end{figure}

\begin{figure}[htbp]
  \centering
  \includegraphics[width=1.0\linewidth]{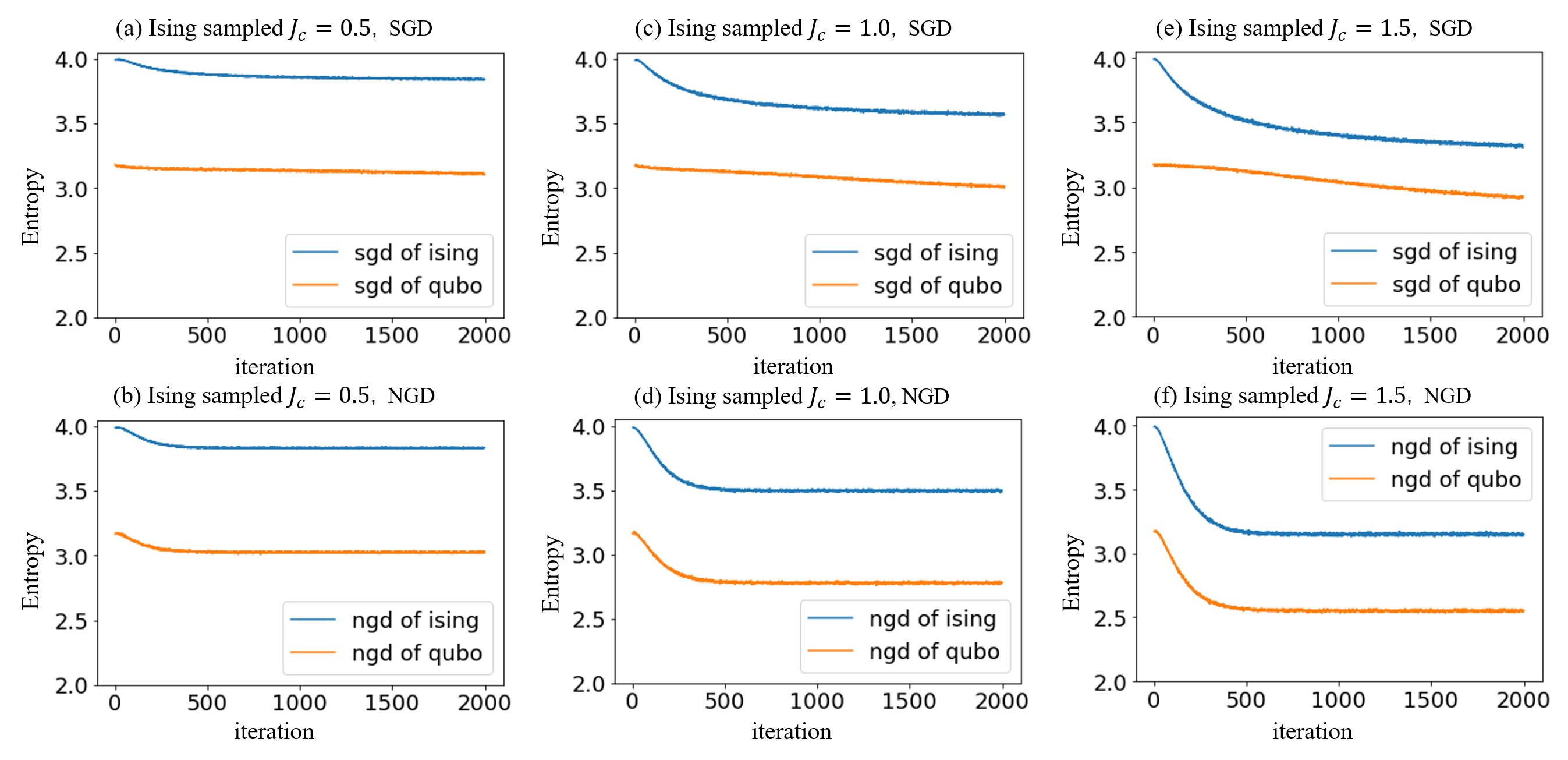}
  \caption{(Color online) Spectral entropy of the FIM on Ising-sampled synthetic data. Panels (a), (c), and (e) show SGD results for $J_c=0.5$, $1.0$, and $1.5$, respectively, while panels (b), (d), and (f) show the corresponding NGD results. The Ising encoding consistently has larger spectral entropy than the QUBO encoding, indicating that the curvature is distributed more evenly across parameter directions. In contrast, the lower spectral entropy of QUBO reflects a more concentrated and anisotropic FIM spectrum. This supports the interpretation that QUBO has more ill-conditioned learning geometry, which slows SGD convergence.}
  \label{fig:entropies_ising}
\end{figure}

\begin{figure}[htbp]
  \centering
  \includegraphics[width=1.0\linewidth]{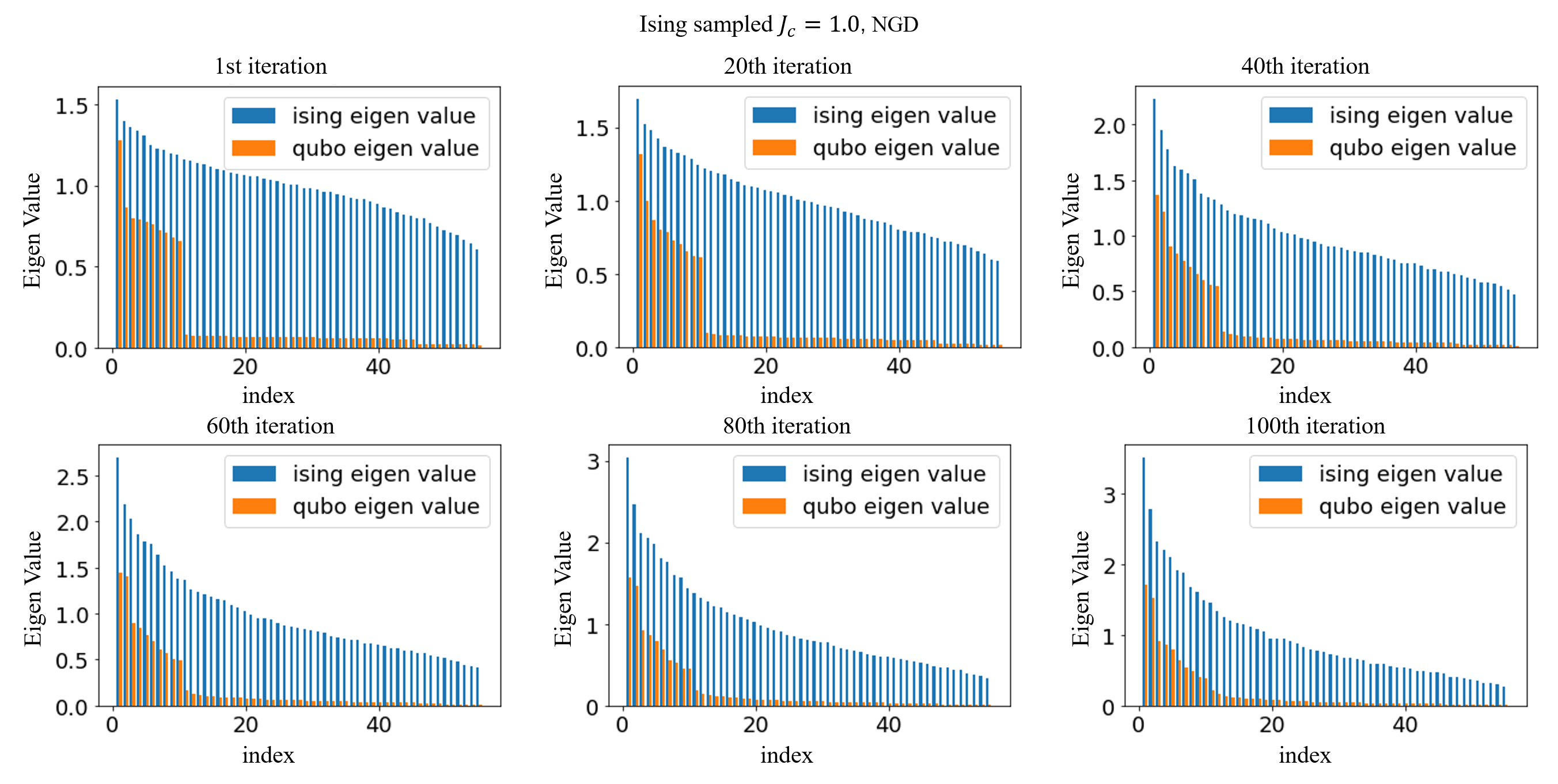}
  \caption{(Color online) Evolution of FIM eigenvalues on Ising-sampled data with $J_c=1.0$ under NGD. The eigenvalues are sorted in descending order at each plotted iteration. Across the early training stage, the Ising FIM exhibits a broader spectrum with many non-negligible eigenvalues, whereas the QUBO FIM shows a sharp drop after a small number of large eigenvalues. This indicates that QUBO has many small-curvature directions and a more anisotropic FIM structure. The spectral difference explains why SGD is sensitive to the choice of variable encoding, while NGD can compensate for this anisotropy by preconditioning the gradient with the FIM metric.}
  \label{fig:eigenvals_ising_ngd}
\end{figure}

\begin{figure}[htbp]
  \centering
  \includegraphics[width=1.0\linewidth]{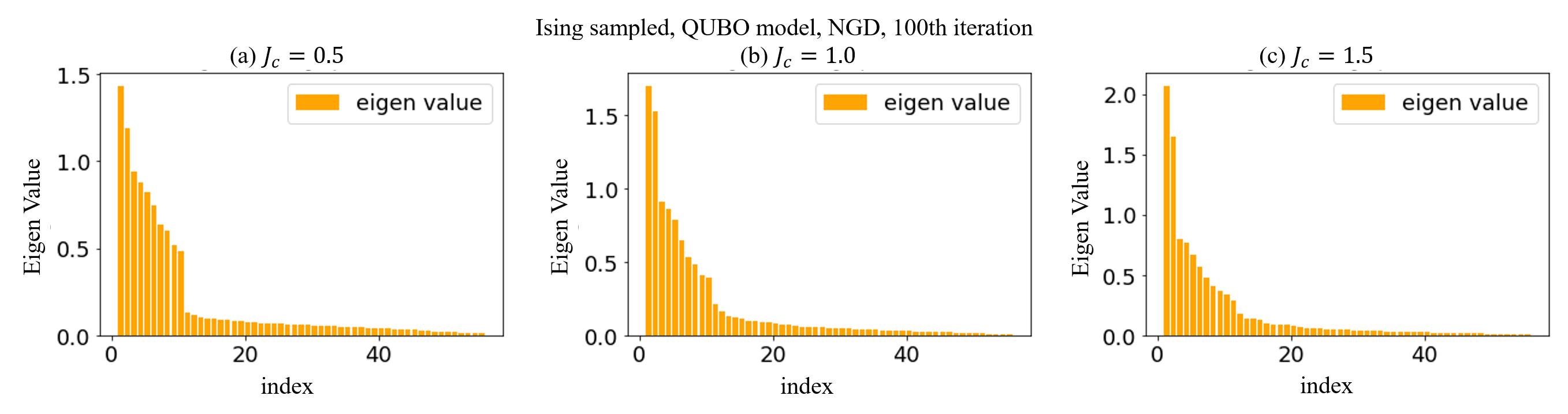}
  \caption{(Color online) FIM eigenvalues for the QUBO encoding at the 100th iteration under NGD for different coupling scales. Panels (a), (b), and (c) correspond to $J_c=0.5$, $1.0$, and $1.5$, respectively. In all cases, the QUBO FIM has only a small number of large eigenvalues and many near-zero eigenvalues. This separation indicates strong curvature anisotropy and the presence of many weakly constrained parameter directions. The persistence of very small eigenvalues is particularly important because these directions can slow SGD updates and make the learning dynamics more sensitive to the representation.}
  \label{fig:eigenval_ising_qubo}
\end{figure}

\begin{figure}[htbp]
  \centering
  \includegraphics[width=1.0\linewidth]{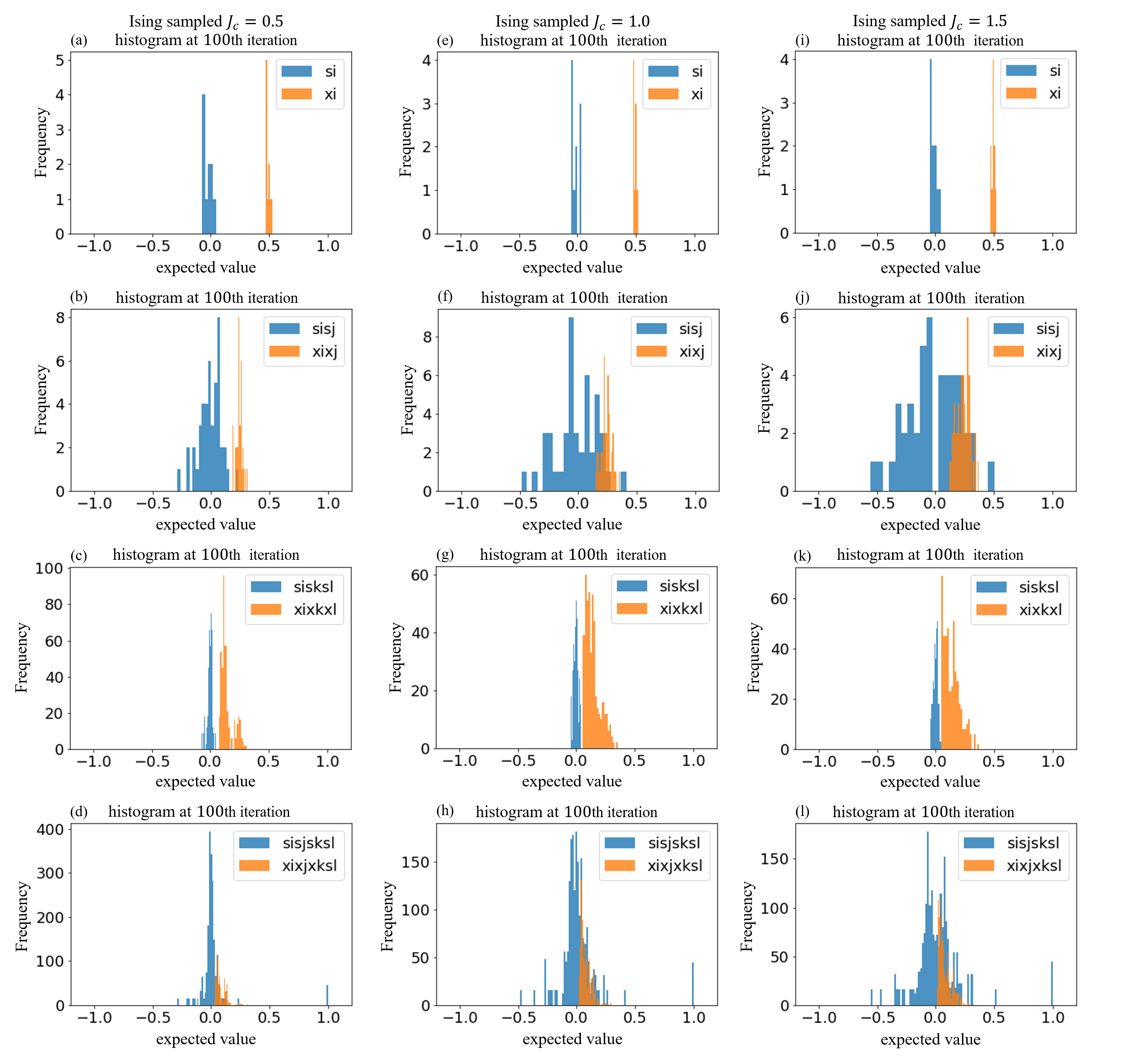}
  \caption{(Color online) Histograms of model expectations for Ising and QUBO sufficient statistics at the 100th iteration under NGD. The columns correspond to $J_c=0.5$, $1.0$, and $1.5$, respectively. Panels (a), (e), and (i) compare first-order statistics, panels (b), (f), and (j) compare second-order statistics, panels (c), (g), and (k) compare third-order statistics, and panels (d), (h), and (l) compare fourth-order statistics. The Ising statistics are centered around zero because the variables take values in $\{-1,+1\}$, whereas the QUBO statistics are non-negative because the variables take values in $\{0,1\}$. In particular, the QUBO first- and second-order moments concentrate around typical positive values such as $E[x_i]\simeq 0.5$ and $E[x_i x_j]\simeq 0.25$. This non-centered structure produces stronger cross-correlations among sufficient statistics and contributes to the off-diagonal FIM blocks observed in the QUBO representation.}
  \label{fig:hists_ising}
\end{figure}

Fig. \ref{fig:kls_ising} further examines the learning trajectories on Ising-sampled synthetic data for different coupling scales. Consistent with the BAS results, SGD exhibits a clear encoding-dependent gap, whereas NGD strongly suppresses this gap. Fig. \ref{fig:entropies_ising} shows that the Ising encoding has larger spectral entropy than the QUBO encoding throughout training, indicating a more evenly distributed curvature spectrum. The eigenvalue spectra in Figs. \ref{fig:eigenvals_ising_ngd} and \ref{fig:eigenval_ising_qubo} support this interpretation by showing that QUBO has only a few large eigenvalues and many near-zero directions. Finally, Fig. \ref{fig:hists_ising} visualizes the underlying moment distributions and shows that the non-centered QUBO statistics induce stronger cross-correlations among sufficient statistics.

In QUBO, the FIM eigenvalues fall off sharply after the 11th component, and a smaller $ J_c$ extends the period during which very small eigenvalues persist. For Ising, the first‑order moments $ \mathbb{E}[s_i] $ and third‑order moments $ \mathbb{E}[s_{i}s_{k}s_{l}] $ are distributed around zero, while second‑ and fourth‑order moments $ \mathbb{E}[s_{i}s_{j}] $, $ \mathbb{E}[s_{i}s_{j}s_{k}s_{l}] $ are symmetric around zero. In contrast, QUBO’s $ \boldsymbol{x} \in \{0,1\} $ asymmetry yields non‑negative moment distributions; in particular $ \mathbb{E}[x_i] \approx 0.5 $ and $ \mathbb{E}[x_ix_j] \approx 0.25 $. Consequently, in Ising the off‑diagonal FIM block $ F_{h,J} = \mathbb{E}[s_is_ks_l] - \mathbb{E}[s_i]\mathbb{E}[s_ks_l] $ is very small and the FIM is approximately block‑diagonal, whereas in QUBO the off‑diagonal block is non‑zero.

Writing the FIM in block form for first- and second-order parameters,
\begin{equation}
F_{\theta} =
\begin{pmatrix}
F_{11} & F_{12} \\
F_{21} & F_{22}
\end{pmatrix},
\label{eq:block-fim}
\end{equation}
The Schur complement inequality bounds the minimum eigenvalue as \cite{horn2012matrix, smith1992some}
\begin{equation}
\lambda_{\min}(F_\theta)\le \lambda_{\min}(F_{22}-F_{21}F_{11}^{-1}F_{12}).
\end{equation}
Since $ F_{12}=F_{21} \approx 0 $ for Ising while it is non-zero for QUBO, the latter tends to produce smaller eigenvalues.

\subsection{Effect of diagonal-FIM approximation}
\begin{figure}[htbp]
  \centering
  \includegraphics[width=1.0\linewidth]{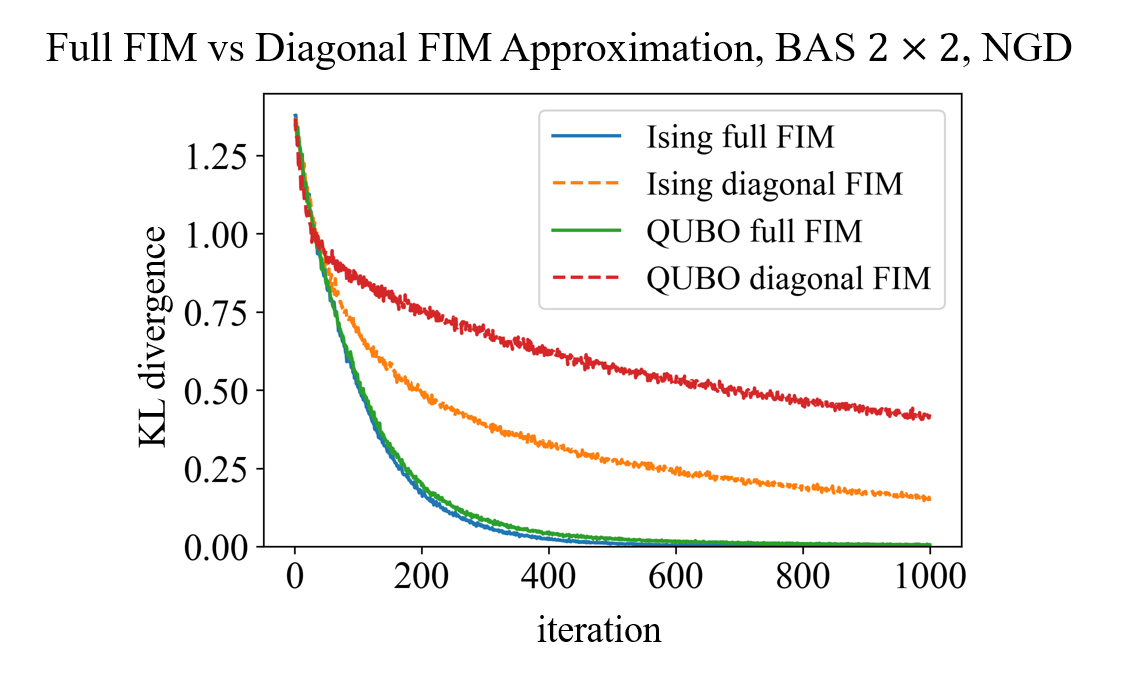}
  \caption{(Color online) Effect of diagonal FIM approximation on NGD for BAS $2\times2$. The full-FIM and diagonal-FIM NGD experiments were performed using the same learning rate, $\eta_{\mathrm{NGD}}=0.01$, and the same damping parameter, $\lambda=10^{-3}$. Full-FIM NGD gives nearly identical and rapid convergence for the Ising and QUBO encodings. When only the diagonal part of the FIM is retained, convergence becomes slower for both encodings, and the degradation is more pronounced for QUBO. This result shows that approximate NGD can exhibit behavior different from exact NGD, and that off-diagonal curvature information is particularly important for QUBO, where cross-correlations among sufficient statistics are stronger.}
  \label{fig:diagonalfim}
\end{figure}

We further examined how this off-diagonal structure affects approximate NGD.
Fig. \ref{fig:diagonalfim} compares full-FIM NGD and diagonal-FIM NGD on BAS $2\times2$. To isolate the effect of the FIM approximation, the full-FIM and diagonal-FIM experiments were performed with the same learning rate $\eta_{\mathrm{NGD}}=0.01$ and the same damping parameter $\lambda=10^{-3}$. With the full FIM, Ising and QUBO exhibit almost identical and rapid
convergence. However, when only the diagonal part of the FIM is retained, convergence becomes slower for both encodings, and the degradation is more pronounced for QUBO. This indicates that approximate NGD can exhibit behavior different from exact NGD, especially when the approximation discards off-diagonal curvature information. Since QUBO has stronger cross-correlations among sufficient statistics, diagonal approximation is less effective for QUBO than for Ising. This result suggests that scalable NGD methods for QUBO should preserve important off-diagonal or block-structured curvature information, for example through block-diagonal or low-rank FIM approximations.

\subsection{RBM experiments}

We tested whether the same qualitative behavior appears in a sparse architecture using an RBM.

\begin{figure}[htbp]
  \centering
  \includegraphics[width=1.0\linewidth]{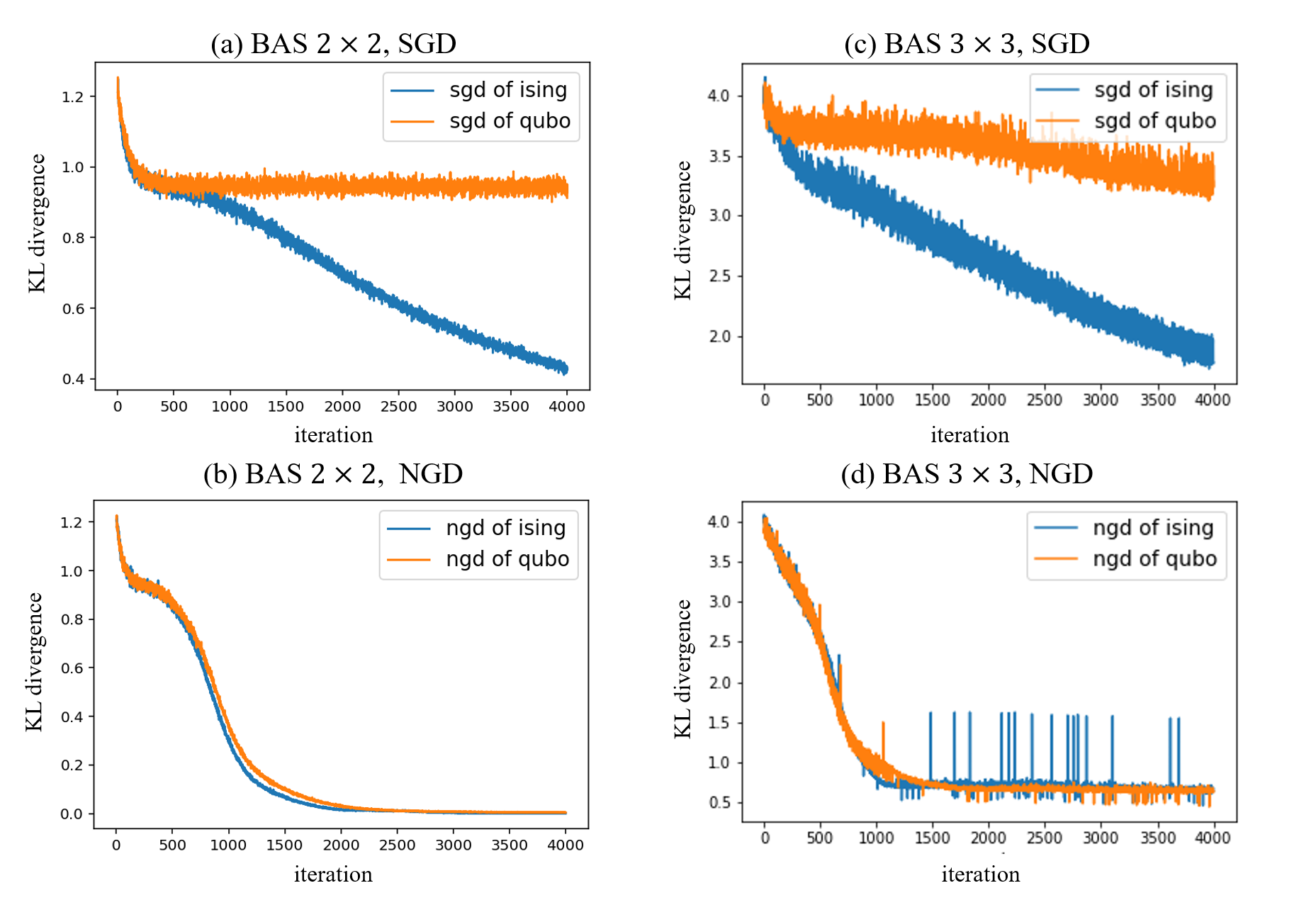}
  \caption{(Color online) Learning trajectories on BAS datasets using an RBM with $n_h=4$ hidden units. Panels (a) and (c) show SGD results for BAS $2\times2$ and BAS $3\times3$, respectively, while panels (b) and (d) show the corresponding NGD results. The numbers of training samples were $450$ and $1{,}120$ for BAS $2\times2$ and BAS $3\times3$, respectively. The same qualitative tendency as in the fully connected BM setting is observed: under SGD, the Ising encoding converges faster than the QUBO encoding, whereas NGD substantially reduces the encoding-dependent gap. Although stochastic fluctuations are visible in the BAS $3\times3$ NGD case, the results suggest that the representation-dependent learning behavior is not exclusive to the fully connected BM setting.}
  \label{fig:kls_bas_rbm}
\end{figure}

Fig. \ref{fig:kls_bas_rbm} shows the KL-divergence trajectories for RBM with $n_h=4$ hidden units on BAS $2\times2$ and BAS $3\times3$. Under SGD, the Ising encoding converges faster than the QUBO encoding for both datasets, consistent with the fully connected BM results. Under NGD, the gap between the two encodings is substantially
reduced, although stochastic fluctuations are visible in the BAS $3\times3$ case. These results suggest that the representation-dependent learning behavior observed in this study is not exclusive to the fully connected BM setting, but can also appear in RBM-type sparse architectures. A systematic study of larger and more diverse sparse structures remains an important
direction for future work.

\section{Conclusion}

We presented a unified analysis of BM learning dynamics from an information-geometric perspective, comparing Ising $\{-1,+1\}$ and QUBO $\{0,1\}$ variable encodings while controlling the sampler, optimizer, and learning-rate design. The results show that the Ising encoding consistently achieves faster KL-divergence reduction under SGD, whereas NGD largely removes the encoding-dependent gap by incorporating the FIM metric.

The main findings are summarized as follows:
\begin{itemize}
  \item \textbf{Stochastic Gradient Descent (SGD):}  
  The Ising encoding consistently achieves faster Kullback–Leibler (KL) divergence reduction and requires fewer iterations to converge than QUBO. This advantage arises from better numerical conditioning, as QUBO introduces parameterization-dependent gradient scales and cross-correlations that increase curvature anisotropy.
  
  \item \textbf{Natural Gradient Descent (NGD):}  
  When curvature is properly accounted for via the full Fisher information matrix (FIM) metric, convergence behavior becomes nearly identical across encodings. This demonstrates that NGD effectively restores representation invariance by rescaling updates according to local geometry.
  
  \item \textbf{Fisher Information Spectra:}  
  During training, the FIM exhibits (a) strong anisotropy in early iterations, (b) progressive redistribution of spectral mass, and (c) systematically smaller eigenvalues and lower spectral entropy for QUBO. These characteristics originate from persistent cross-block correlations between first- and second-order sufficient statistics.
\end{itemize}

The fixed-learning-rate experiment and the SGD FIM spectra show that the advantage of Ising is not simply caused by an unfair learning-rate choice against QUBO. The diagonal-FIM experiment also demonstrates that approximate NGD can reintroduce representation-dependent
differences, especially for QUBO, when important off-diagonal curvature information is discarded. Finally, RBM results indicate that the observed behavior is not limited to the fully connected BM setting.
The Ising encoding provides more isotropic curvature and better-conditioned learning under SGD, leading to faster convergence. NGD further enhances both stability and speed for either encoding, though it requires computing or approximating \( F^{-1} \), whose cost scales with parameter dimensionality. Efficient second-order approximations—such as block-diagonal or low-rank FIMs, Kronecker-factored curvature (K-FAC), or stochastic inverse estimators like Hutchinson, Lanczos, Shampoo, and AdaHessian methods~\cite{martens2015optimizing, hutchinson1989stochastic, gupta2018shampoo, yao2021adahessian, ubaru2017fast}—offer promising routes to scalable NGD implementations.

Because the FIM of a BM equals the covariance of sufficient
statistics, initialization and encoding-level preprocessing---in particular, centering and scaling of variables or sufficient statistics---can align first- and second-order moments and substantially reduce training iterations,
especially for QUBO variables.
Future research directions include:  
(i) principled moment-matching initialization schemes,  
(ii) adaptive damping policies informed by spectral monitoring,  
(iii) hardware-aware studies using quantum and digital annealers—including online calibration of the effective inverse temperature for quantum annealing–based RBM training~\cite{goto2025online}, and  
(iv) extensions to deep energy-based models where variable representation interacts with hierarchical structure, as suggested by three-layer RBM kernel learning and the three-layer QDRBM~\cite{hasegawa2026kernel, yonaga2025quantum}.

\begin{acknowledgment}
We received financial support from the Cabinet Office programs for bridging the gap between R\&D and IDeal society (Society 5.0), Generating Economic and Social Value (BRIDGE), and the Cross-ministerial Strategic Innovation Promotion Program (SIP) (No.~23836436).
\end{acknowledgment}
\bibliographystyle{jpsj}
\bibliography{library}

\end{document}